\documentclass{article}

\usepackage{microtype}
\usepackage{graphicx}
\usepackage{subfigure}
\usepackage{booktabs} % for professional tables

\usepackage{hyperref}

\usepackage[accepted]{icml2023}

\usepackage{amsmath}
\usepackage{amssymb}
\usepackage{mathtools}
\usepackage{amsthm}

\usepackage{xspace}
\newcommand{\our}{GWL\xspace}
\newcommand{\outputdist}{output distribution\xspace}

\usepackage{enumitem}

\usepackage[capitalize,noabbrev]{cleveref}

\theoremstyle{plain}

\theoremstyle{definition}

\theoremstyle{remark}

\usepackage[textsize=tiny]{todonotes}

\begin{document}

\twocolumn[
\icmltitle{Gradient-based Wang--Landau Algorithm: A Novel Sampler for \\Output Distribution of Neural Networks over the Input Space}




\begin{icmlauthorlist}
\icmlauthor{Weitang Liu}{yyy}%{equal,yyy}
\icmlauthor{Ying-Wai Li}{comp}
\icmlauthor{Yi-Zhuang You}{sch}
\icmlauthor{Jingbo Shang}{yyy}
\end{icmlauthorlist}

\icmlaffiliation{yyy}{Department of Computer Science Engineering, University of California, San Diego, La Jolla, USA}
\icmlaffiliation{comp}{Computer, Computational, and Statistical Sciences Division
Los Alamos National Laboratory, USA}
\icmlaffiliation{sch}{Department of Physics, University of California, San Diego, La Jolla, USA}

\icmlcorrespondingauthor{Weitang Liu}{wel022@ucsd.edu}

\icmlkeywords{Machine Learning, ICML}

\vskip 0.3in
]



\printAffiliationsAndNotice{}  % leave blank if no need to mention equal contribution

\begin{abstract}
The \outputdist of a neural network (NN) over the \emph{entire input space} captures the complete input-output mapping relationship, offering insights toward a more comprehensive NN understanding.
Exhaustive enumeration or traditional Monte Carlo methods for the entire input space can exhibit impractical sampling time, especially for high-dimensional inputs. 
To make such difficult sampling computationally feasible, in this paper, we propose a novel Gradient-based Wang-Landau (\our) sampler. 
We first draw the connection between the \outputdist of a NN and the density of states (DOS) of a physical system.
Then, we renovate the classic sampler for the DOS problem, Wang--Landau algorithm, by replacing its random proposals with gradient-based Monte Carlo proposals.
This way, our \our sampler investigates the under-explored subsets of the input space much more efficiently.
Extensive experiments have verified the accuracy of the \outputdist generated by \our and also showcased several interesting findings --- for example, in a binary image classification task, both CNN and ResNet mapped the majority of human unrecognizable images to very negative logit values. 
%\yingwai{Acronyms should be defined upon first use.}

\end{abstract}

\section{Introduction}

\begin{figure}[t]
    \centering
    \includegraphics[width=\linewidth]{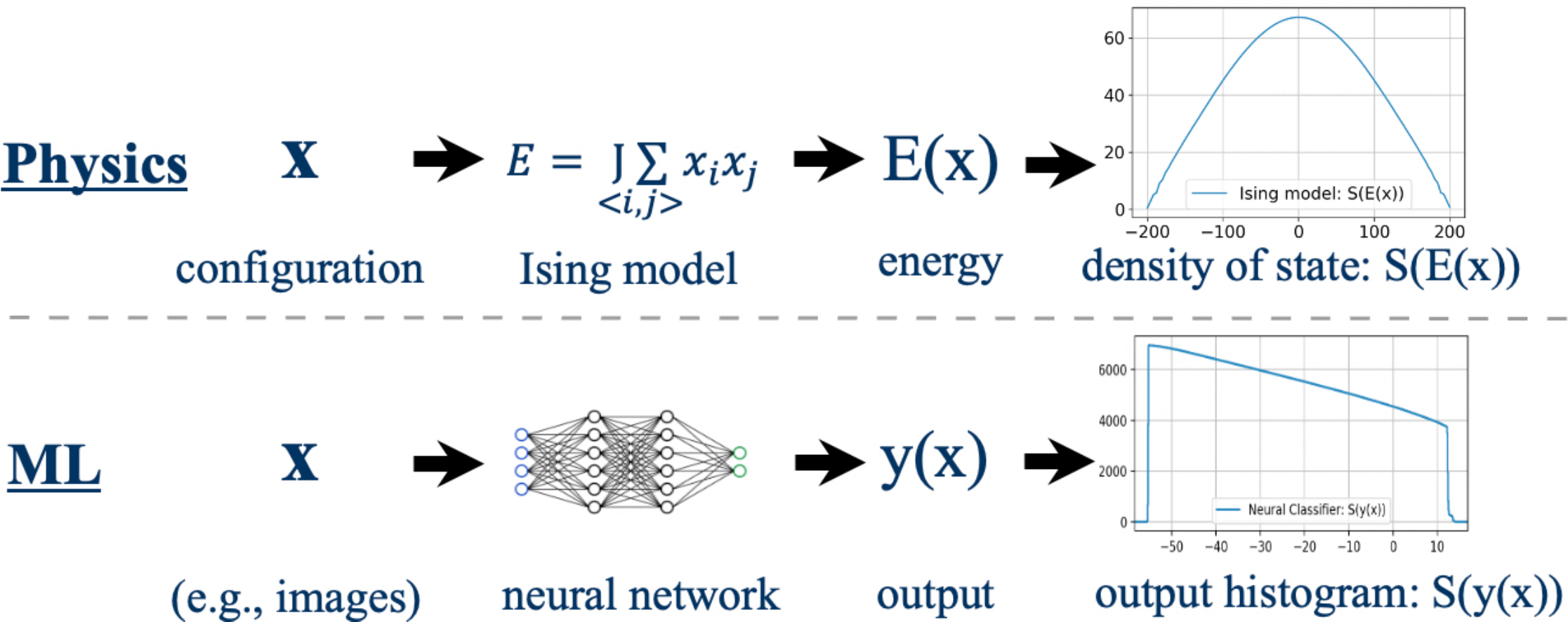}
    \vspace{-5mm}
    \caption{The energy density of states (DOS) of a physical system \emph{vs.} the output distribution of a deep neural network.
    % \jingbo{revise this caption later. rotate this plot to make it wider but shorter.}
    }
    \vspace{-3mm}
    \label{fig:connection}
\end{figure}

\begin{figure*}[t]
     \centering
     \subfigure[Different types of input samples]{
         \centering
         \includegraphics[width=0.55\linewidth]{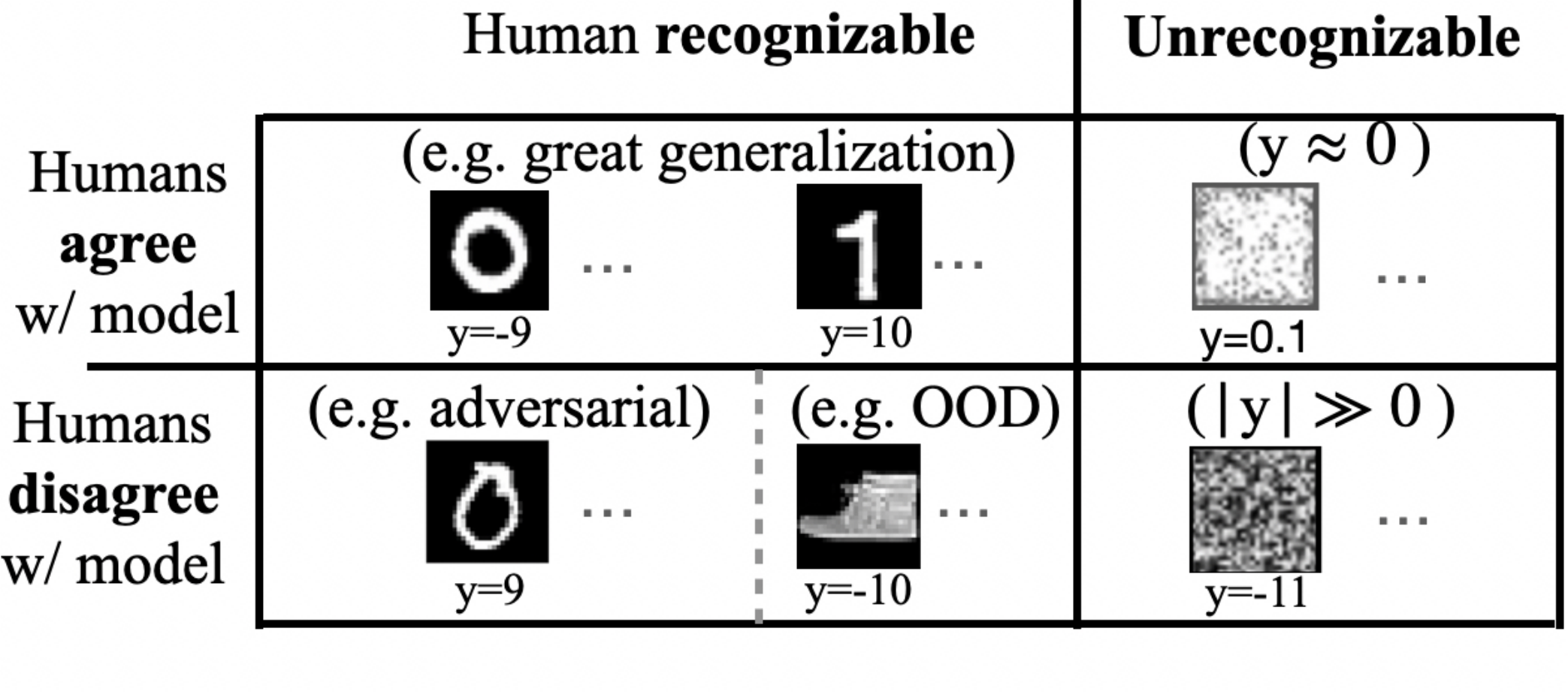}
         \label{fig:intro1}
     }
     \subfigure[The \outputdist of an example binary classifier]{
         \centering
         \includegraphics[width=0.4\linewidth]{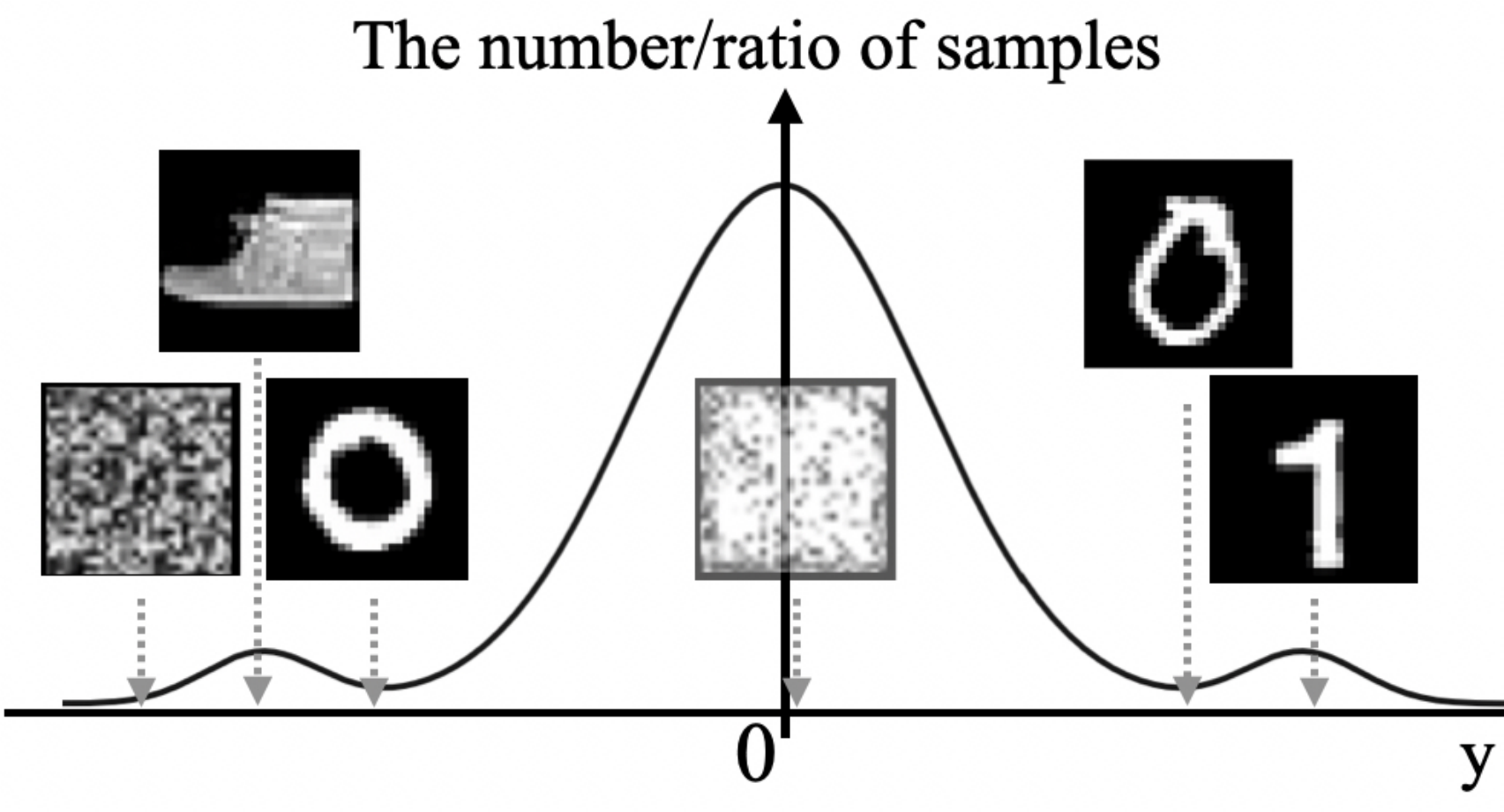}
         \label{fig:intro2}
     }
    \vspace{-3mm}
    \caption{Input types and the example \outputdist for binary classification between digits 0 and 1. The entire input space covers all possible gray-scale images of the same shape. $y$ is the output (logit) with respect to input $\mathbf{x}$.}
    \vspace{-5mm}
\end{figure*}

The input-output mapping relationship of a trained neural network (NN) is the key to understand a trained NN.
Existing works measure the accuracy of a NN based on such mapping relations over 
(pre-defined) \emph{subsets} of the input space, such as 
% by measuring the accuracy on 
in-distribution subsets~\cite{dosovitskiy2020vit, tolstikhin2021mixer, steiner2021augreg,chen2021outperform,zhuang2022gsam, He2015}, 
out-of-distribution (OOD) subsets~\citep{liu2020energy,hendrycks2016baseline,hendrycks2018deep,hsu2020generalized,lee2017training,lee2018simple},
and adversarial subsets~\cite{szegedy2013intriguing,rozsa2016adversarial,miyato2018virtual,kurakin2016adversarial}.

Given the recent trend of applying NNs to open-world, non-IID applications~\citep{cao2022openworld, sun2022open},
we argue that it is crucial to obtain the complete \emph{\outputdist} of a trained NN over the \emph{entire input space}.
This \outputdist can offer a complete picture about the number of inputs mapped to certain output values.
Note that the entire input space here includes all kinds of inputs mentioned above and even \emph{human unrecognizable} inputs (see Figure~\ref{fig:intro1}). 
As a pilot study, we focus on binary classification --- given a trained binary NN classifier, we aim to sample the entire input space to obtain the \outputdist, i.e., a histogram that counts the number of input samples mapped to certain logit values, as shown in Fig~\ref{fig:intro2}.
The sampling procedure would also offer more fine-grained information as side products, such as representative input samples corresponding to a certain range of output values.

A straightforward solution is exhaustive enumeration or traditional Monte Carlo methods~\citep{chen2014stochastic, welling2011bayesian, li2016preconditioned, xu2018global}. However, the sampling time would become impractical, or the sampler could get stuck in a subset of input space, especially for high-dimensional inputs. 
% do not sample based on uniform measure as required and thus need multiple complete sampling process in order to sample based on uniform measure (Sec.~\ref{sec:wl}).
To overcome these issues, in this paper, we propose a novel sampler called Gradient-based Wang--Landau (\our) sampling as follows. 

We first connect the \outputdist of a NN to the \emph{density of states} (DOS) of a physical system through an analogy between the system energy and neural network output, as shown in Figure~\ref{fig:connection}.
From the physics point of view, the input $\mathbf{x}$ to the neural network can be viewed as the configuration $\mathbf{x}$ of the system; 
the neural network output (e.g., logit values in binary classifier) $y(\mathbf{x})$ corresponds to the energy function $E(\mathbf{x})$; 
the \outputdist of a NN is then analogous to the DOS of a physical system, which is the number of configurations corresponding to the same energy value. 
The log scale of the DOS is the microcanonical entropy associated with the energy, $S(E(\mathbf{x}))$.

Our new sampler \our is a novel renovation of the classic sampler for the DOS problem, Wang--Landau algorithm~\citep{wang2001efficient}, where we  replace its random proposals with gradient-based Monte Carlo proposals.
Given the overwhelming number of human unrecognizable inputs in the entire input space, if one adopts the traditional Monte Carlo proposal in the Wang--Landau algorithm, i.e., by changing pixel values at random, the sampling process is likely to get stuck in this human unrecognizable subset. 
Thus, we propose to apply a gradient-based proposal following Gibbs-with-Gradients~\citep{grathwohl2021oops}, which proves to be efficient to propose in-distribution inputs for a trained NN model.
This way, our \our sampler investigates the under-sampled subsets of the input space much more efficiently.
The accuracy of \our has been empirically verified on a small toy dataset --- the \outputdist generated by \our aligns perfectly with the result of exhaustive enumeration. 

More importantly, by analyzing the \outputdist generated by \our, we showcase several interesting findings of CNN and ResNet in a binary classification task based on real-world pictures.
First, our experiments show that in both CNN and ResNet, the dominant output values are very negative and the vast majority of them correspond to human-unrecognizable input images. 
This supplies direct evidence to the well-known overconfidence issue in NNs~\citep{nguyen2015deep}.
% indicates that the NN models map an overwhelmingly large number of unrecognizable images to the exceptionally high probabilities(overconfident prediction probabilities) 
% \yingwai{Do you mean ``the model maps unrecognizable images to a prediction overconfidently''?}. 
Second, when we focus on the output values where the in-distribution inputs correspond to, human-unrecognizable inputs still dominate significantly. 
This result presents significant challenges to the out-of-distribution (OOD) detection problems. 
Third, we observe a clear background darkness pattern of the representative samples of CNN and ResNet when the output logit value increases, and speculate these models simply utilize such ``backdoors'' to predict the labels of the digits without truly understanding the semantics of the images. 

%Second, we can derive the relative difference between the dominant peak of output values and the other output values, especially those where the in-distribution inputs correspond to. \jingbo{what does the ratio mean here? not clear}

In summary, we demonstrate that sampling the entire input space to obtain the \outputdist of a trained NN is computationally feasible, and it can provide new and interesting insights for future systematic investigation. 
Our contributions are summarized as follows.
\begin{itemize}[nosep,leftmargin=*]
    \item We tackle the challenging yet important problem to uncover the \outputdist of a NN over the entire input space. 
    Such \outputdist offers a novel perspective to understand NNs.
    \item We connect this \outputdist to the DOS in physics and successfully renovate the Wang--Landau algorithm using a gradient-based proposal, which is a critical component to sample the entire output space as much as possible, and to improve efficiency.
    \item We conduct extensive experiments on toy and real-world datasets %using CNN and ResNet 
    to confirm the accuracy of our proposed sampler.
    \item \our sampler allows for detailed investigation of the input-output mapping of NNs, facilitating further studies systematically.
\end{itemize}

%\noindent\textbf{Reproducibility.} We will release our code on GitHub\footnote{\url{https://github.com/anonymous_repo}}.

\section{Problem definition}

%\jingbo{I think this 3.1 section is more like a problem definition instead of a part of the proposed method. }

%\jingbo{and aims to obtain xxx. just reiterate the goal briefly here}. 
In the traditional setting, binary neural classifiers model the class distribution through logit $z$. A neural classifier parameterized by $\theta$ learns $p_\theta(z|\mathbf{x})=\delta(z-y_\theta(\mathbf{x}))$ through a function  $y_{\theta}:\mathbf{x}  \rightarrow z \in \mathbb{R}$, where $\mathbf{x} \in \Omega$, $\Omega \subseteq \{0,...,N\}^D$ for images, and $\delta$ is the Dirac delta function. $\Omega$ aligns with Gibbs-With-Gradient's setting to be discrete.  %\jingbo{it reads like x is discrete? it can be continuous too right? also, we are working on the entire input space. Shall we define it as $\Omega$?} 
% \wei{the current setting does not work in continuous x space because of GWG's setting. Also we cannot enumerate if x is continuous.} 
%to two classes is simply the output without taking any activation function $y_{\theta}:\mathbf{x}  \rightarrow z \in \mathbb{R}$. 
%\jingbo{we are using $f()$ here but in intro it is $y()$. I don't have a preference but wish to be consistent.}
% The logit then inputs to a sigmoid activation function to %\jingbo{this ``to'' reads a bit weird to me. Double check the grammar.} 
% produce a prediction probability of the positive class. %\jingbo{better say positive class here.}. 
% %, where $f_\theta(\mathbf{x})$ is the logit for class $y\in \mathbb{R}$. 
% This mapping from $\mathbf{x}$ to class $c$ defines a conditional relationship:
% \[
% p_\theta(c=1|\mathbf{x})= \frac{1}{1+e^{-y_\theta(\mathbf{x})}}.
% \]
% As $p_\theta(c=0|\mathbf{x})=1-p_\theta(c=1|\mathbf{x})$, $p_\theta(c|\mathbf{x})$ is a one-to-one mapping from the logit $y_\theta(\mathbf{x})$ and thus there is a meaningful distribution $p_\theta(z|\mathbf{x})$. The samples with more positive logit values correspond to class $1$ and vice versa for class $0$. Thus, we can simply focus on the logit distribution $p_\theta(z|\mathbf{x})$. 

The above model does not define the distribution of the data $\mathbf{x}$. %\jingbo{is it better to say over the input space? I feel there should be multiple items when we using the word ``over''?}.
This work aims to obtain the output value distribution of binary classifiers in the entire input space: $\Omega = \{0,...,N\}^D$.
% from Prof. Ma...
Here we assume that the input %\wei{do we really need the measure?} %\jingbo{this ``it'' is not clear. I understand that it refers to the data distribution of $\mathbf{x}$. Better to make it explicit to avoid ambiguity.} 
follows a uniform distribution $\mu(\mathbf{x})$ over the domain $\Omega$ of $\mathbf{x}$. %\yingwai{The notation is a bit redundant to me. If $p(x) = \mu$, just use $\mu(x)$ instead of $p(x)$.}
We define the joint distribution 
\[
p_\theta(z,\mathbf{x}) = p_\theta(z|\mathbf{x}) \mu(\mathbf{x}).
\]

Our goal is to obtain the logit (output) distribution $p_\theta(z)$, %\jingbo{this ``ultimately'' is a bit odd. It's better to position it as ``One can easily marginalize the .... to define the density over the label space.'' Our goal is only the logit output distribution.}
which can be obtained by marginalizing the joint distribution over the input space $\Omega$:
\[
p_\theta(z) = \sum_{\Omega} p_\theta(z|\mathbf{x}) \mu(\mathbf{x}).
%=\int_{\Omega} \frac{1}{1+e^{-y_\theta(\mathbf{x})}} \mu(d\mathbf{x}).
\]
To sample from the distribution $p_\theta(z)$, we can first sample $\mathbf{x}_i \sim \text{Uniform}(\Omega)$, then condition on the sampled $\mathbf{x}_i$ to obtain $z_i \sim p_\theta(z|\mathbf{x}_i)$. While a uniform sampler in principle can solve this problem, 
%\jingbo{we'd better not say the reason as the 2nd half here. From the exp results, we already learned that it is possible that the output values of in-distribution samples can be largely overlapped with those popular output values too. So just say that uniform sampling can take almost forever to converge. No one will challenge this argument.}
it can take an impractically long time to converge. %sample the output values to which in-distribution inputs correspond for a trained model which should memorize the training structure.

\section{Method}
In this section, we discuss the connection between our problem to the density of states (DOS), introduce both Wang-Landau algorithm and the Gibbs-With-Gradient proposal method as a background, and present our new sampler Gradient-Wang-Landau (GWL) algorithm. 
%\jingbo{write one or two sentences here to give an overview of this section.}

\subsection{Connection to Density of States in Physics}

%\yzy{should we use $\mathcal{E}$ for the mapping, so as to avoid notation like Ene, also what does E correspond to in the original problem setup?}
In statistical physics, given the energy function $E: \mathbf{x} \rightarrow \mathcal{E} \in \mathbb{R}$ %\jingbo{probably define what is an energy function? like it is mapping a vector to a scalar?}
, the DOS $\rho(\mathcal{E})$ is defined as %\jingbo{what is E? a scalar variable?}
\[
\rho(\mathcal{E}) = \sum_{\mathbf{x} \in \Omega} \delta(\mathcal{E}-E(\mathbf{x})),
\]
%\jingbo{Is $\delta$ sigmoid function? define it.}
where $\delta$ is the Dirac delta function and $\Omega$ is the domain of $\mathbf{x}$ where $\mathbf{x}$ is valid. 
The DOS can be viewed as a probability distribution in the energy space; its log-probability defines the entropy $S$:
% \[
% \rho(E) \propto \frac{\exp(S(E))}{W}
% \]
% Where $W$ is a normalization constant $W=\sum_E \exp(S(E))$. \wei{here it seems the above $\rho$ are defined equivalently. I feel like we might need an empirical term $S'$ to emphasize $S'$ is the approximate $S$. When $S'->S$, then the uniform distribution assumptio below holds.}
\[
%\rho(\mathcal{E}) = \exp(S(\mathcal{E}))
S(\mathcal{E}) = \ln (\rho(\mathcal{E})).
\]
Boltzmann constant is taken to be $1$ in our setting. DOS is meaningful because many physical quantities depend on energy or its integration but not the specific input $\mathbf{x}$.

%\jingbo{just revise this part (copied from intro) to make it more formal and detailed.}
We associate the neural network output distribution to DOS in physics by making an analogy between the system energy $\mathcal{E} = E(\mathbf{x})$ and NN output $z = y(\mathbf{x})$.
This connection is based on the observation that the energy function in physics maps an input configuration to a scalar-valued energy; similarly, a binary neural classifier maps an image to a logit. Both the logit and energy are treated as the direct output of the mapping. Other quantities, such as the loss, are derived from the output. The desired output distribution can be obtained similarly as sampling the DOS in physics, which is the count of the configurations given an energy value. The output distribution and DOS are both defined in the entire input space. 

\subsection{Traditional Samplers Are Not Directly Applicable}
%\yingwai{WL is a Monte Carlo algorithm...}
\label{sec:mcmc}

Traditional Monte Carlo (MC) samplers~\citep{chen2014stochastic, welling2011bayesian, li2016preconditioned, xu2018global}, in principle, could be applied to sample the output distribution, but they would not be efficient to our study. This is because these algorithms bias the sampler to the more probable domain based on importance sampling. Consequently, a major drawback is that the sampler is easily ``stuck'' in some localized distributions as it is hard for the sampler to overcome the barriers to visit all the possible configurations (or input images in the NN case). This limitation is particularly severe when sampling from multi-modal distributions. Our problem setting, however, not only requires the sampler to sample from a multi-modal distribution. More importantly, the target distribution $S$ is \emph{unknown} upfront and the generated samples have to cover the whole output space. 
Using traditional MC samplers, in the best case scenario, would take an unreasonable time to converge. In the more critical but likely scenario, there is a high risk of obtaining samples that do not truly represent the underlying distribution. 

\subsection{Wang-Landau algorithm and Gibbs-With-Gradient~\label{sec:wl}}
%We briefly introduce the Wang-Landau algorithm and Gibbs-With-Gradient (GWG) as background/preliminary for our new sampler of comprehensive understanding of the target model. 
%\jingbo{To make this paper self-contained, we briefly introduce the Wang-Landau algorithm and Gibbs-With-Gradient (GWG) here as background/preliminary. Add one sentence like this here.}

\textbf{Wang--Landau (WL) algorithm} %is a Markov chain Monte Carlo 
was originally designed to determine the DOS $\rho(\mathcal{E})$ of a physical system~\citep{wang2001efficient}, when the DOS is not known \emph{a priori} and would be determined on-the-fly. It is therefore a suitable tool for estimating the true distribution of our NN output as it is also unknown before the sampling. 
WL uses a histogram to store the instantaneous estimation $\tilde{S}$. WL improves the sampling efficiency by using the inverted distribution as the sampling weight $w(\mathbf{x})$:
\[
w(\mathbf{x}) \propto \exp(-\tilde{S}(E(\mathbf{x}))).
\]
%where $Z$ differs from $W$ as $Z=\sum_\mathbf{x} \exp(-S'(E(\mathbf{x})))$. 
%\yingwai{Note that WL does not sample $p(x)$!} %\wei{ok, marked and will discussed later}
The instantaneous entropy $\tilde{S}$ is updated iteratively until convergence. At the end of the simulation, when the estimation of the entropy approaches the true value $S(\mathcal{E})$, the sampler would sample the entire output space uniformly. 
%we can get an ensemble of $\mathcal{E}$ via $E(\cdot)$ whose probability distribution is:
%\[
%\pi(\mathcal{E}) =\sum_{\mathbf{x} \sim p(\mathbf{x})} \delta(\mathcal{E}-E(\mathbf{x}))
%\]
%When $\tilde{S}(\mathcal{E})$ approaches $S(\mathcal{E})$, the energy distribution $\pi(\mathcal{E})$ approaches to $\mathcal{E}$-independent constant for all the %accessible energy $\mathcal{E}$. 

%The distribution WL samples the exact distribution Equ~\ref{equ:S} and is independent on the choice of temperature. 
Previous work on the sampling of a complex physics system has shown that with the same number of MC steps, WL was able to successfully produce the correct distribution $S$ when the traditional Metropolis MC sampling fail~\cite{li2012surface}. 
This is because WL can overcome energy barriers by accumulating the counts of visits and uses their inverse as sampling biases, a mechanism that traditional MC samplers are missing. 

%: \wei{here the $\mathbf{x}$ have to be decently large so that it can cover all the accessible E so that following can be true}
% \[
% \pi(E) = \frac{1}{W\times Z}
% \]
% In summary the key idea of Wang-Landau algorithm is that inverted distribution $p(E)$ drives the sampler to explore the energy values whose corresponding entropy $S(E)$ is low. 

\noindent The \textbf{Gibbs-With-Gradients (GWG)} method is used for energy-based models (EBM) by sampling
\[
\log p(\mathbf{x})=f(\mathbf{x}) - \log Z,
\]
where $f(\mathbf{x})$ is the unnormalized log-probability, $Z$ is the partition function, and $\mathbf{x}$ is discrete. %Assume a discrete D-dimensional random variable $\mathbf{x} \in \{0,...,N\}^D$. 
Typical Gibbs sampler iterates every dimension $x_i$ of $\mathbf{x}$, computes the conditional probability $p(x_i|x_1,...x_{i-1}, x_{i+1},...,x_D)$, and samples according to this conditional probability.

When the training data $\mathbf{x}$ are natural images and the EBM learns $\mathbf{x}$ decently well, the traditional Gibbs sampler wastes much of the computation. For example, most pixel-by-pixel iterations over $x_i$ in MNIST dataset will be on the black background. GWG proposes a smart proposal that picks the pixel $x_i$ that is more likely to change, such as the pixels around the edge between the bright and dark region of the digits. %\yzy{Between bright and dark regions} 

% \wei{I didn't understand it well} With this intuition, GWG proposes to use a proposal distribution that balance the forward and reverse proposal. With $\mathbf{x}$ in the Hamming ball $H(\mathbf{x})$, the proposal distribution is:
% \[
% q(\mathbf{x}'|\mathbf{x}) \propto \exp(\frac{f(\mathbf{x}')-f(\mathbf{x})}{T})\mathbf{1}(\mathbf{x}' \in H(\mathbf{x}))
% \]

% where T is parameter and usually set as T=2. The benefit of this setting is the proposal is simply the softmax over $f(\mathbf{x}')-f(\mathbf{x})$ for $\mathbf{x}' \in H(\mathbf{x})$. Computing $f(\mathbf{x}')-f(\mathbf{x})$ is still time consuming and thus they propose to use a Taylor expansion up to the first order derivative to approximate it. The proposal distribution becomes: 
% \[
% q_{\text{grad}}(\mathbf{x}'|\mathbf{x}) \propto \exp(\frac{d(\mathbf{x})}{2})\mathbf{1}(\mathbf{x}' \in H(\mathbf{x}))
% \]
% where $d(x)_{ij}=\nabla_x f(x)_{ij} - x_i^T\nabla_x f(x)_i$, $i$ is the $i$-th dimension of $\mathbf{x}$, and $j \in \{0,1,...,N\}$ is the $j$-th value $x_i$ can change to from its current value.  

\subsection{Wang--Landau with Gradient Proposal}

%\jingbo{I would suggest to put the current background section here as the first part to briefly introduce the W-L algorithm and define necessary notations.}

% \jingbo{And then, the W-L with gradient proposal starts from here.} \wei{The key observation of the equilance between the outputs and energy in the previous paragraph is why it is valid to use this sampler. Can we just start from this paragraph?}
Directly applying WL algorithm with random proposals is insufficient to sample the output space efficiently, because a trained neural model learns a preferred mapping through the loss function. For example, a binary classifier maps the training inputs to either the sufficiently positive or negative logit values, which ideally should correspond to the extremely rare but semantically meaningful inputs. %Thus, random proposal in Wang-Landau algorithm will lead to an overwhelmingly large number of rejections of the proposals when the output values that correspond to the semantically meaningful inputs are underexplored. 
%\jingbo{Fig.~\ref{fig:intro2} only works for the very ideal classifiers. Make it clear here. It's quite different than what we have in experiments.}
After the sampler explores and generates the peak centered at 0 where most random samples correspond to (Fig.~\ref{fig:intro2}), it is almost impossible for the sampler with a random proposal to propose an input with meaningful structure (or even in-distribution inputs) so that the other possible output values are explored. Of course, whether those output values correspond to in-distribution inputs is only confirmable after sampling. In summary, it is extremely difficult for the random proposal in WL algorithm to explore all the possible output values. % proposals that, , . the energy functions in physics are symmetric and relatively simple. Thus, the corresponding inputs generally do not have semantic meaning and the random proposal in Wang-Landau algorithm works for these less-structural inputs. However, 

We therefore propose to use the Wang--Landau algorithm framework but replace the MC proposal with the one in Gibbs-With-Gradients (GWG) sampler. GWG has a gradient proposal that takes advantage of the model's learned weights to propose inputs. In order to sample the distribution of the output prediction through GWG, we define log-probability $f(\mathbf{x})$ as:
\[
f(\mathbf{x}) = S(y(\mathbf{x})),
\]
where $S$ is the count in log scale for the bin corresponding to $y({\mathbf{x}})$. The fixed $f(\cdot)$ in the original GWG is now changing in our sampling process given the input $\mathbf{x}$, since the expression for $S$ is unknown and we can only estimate the output distribution from using WL algorithm. GWG requires the gradient of $f$, but since $S$ is approximated using discrete bins, 
% We use an interpolator to acquire a first-order differentiable interpolation for the discrete histogram of entropy.
we apply a first-order differentiable interpolation for taking the derivative of the discrete histogram of entropy. 
%\jingbo{can we add a simple formula here to show the interpolator? I think it's kind of a solution that we proposed to address the practical challenges? worth mentioning it a bit more. Another idea here is to creat a figure to illustrate the process. It might be more telling.}\wei{I used linear and cubic interpolation. Do most of people in ML know those interpolation? It is as simple as linear interpolation or cubic interpolation. That's it. } \jingbo{say it clearly that the interpolator is linear/cubic.} \wei{I will add it to the experiment section. It seems there are some slight differences of the results for ResNet and CNN.}

Similar to the original WL algorithm, we first initialize two histograms with all of their bins set to 0. %\jingbo{it reads like there are only two bins. I assume you are talking about H and S, and each of them can have many bins.}
One of these histograms is for estimating entropy $S$, and the other histogram, $H$, is a counter of how many times the sampler has visited a specific bin. $H$ is also used for checking if all the bins have been visited roughly equally, i.e., a flatness check. %\jingbo{I won't say H is for the flatness check. To me, H is kind of a counter about how many times that the sampler has visited every region. }. 
We first preset the number of iterations that the sampling will perform, as well as a modification factor $f_m$ that is used to update the estimation of entropy $S$ iteratively. 
At each MC step,%\textbf{step} \jingbo{here ``step'' is not the same as the ``English word step''. Let's define it. I think it's useful to discuss the experimental results too. } 
we interpolate $S$ to get a differentiable interpolation, take the derivative of the negation of $S$ with respect to the output $z$ and then the inputs $\mathbf{x}$ using chain rule. GWG uses this gradient to propose the next input that is likely to have a \emph{lower} entropy and be accepted by the sampler. The newly proposed input sample is then accepted or rejected according the acceptance probability:
%{\small
\[
A(\mathbf{x}\rightarrow \mathbf{x}') = \min (1, e^{S_{\mathbf{\mathbf{x}}}-S_{\mathbf{\mathbf{x'}}}}\frac{ 
q(\mathbf{x}|\mathbf{x'})}{q(\mathbf{x'}|\mathbf{x})}%\frac{dS_{\mathbf{x}'}}{dE_{\mathbf{x}'}}\nabla_{\mathbf{x}'}{E}-\frac{dS_{\mathbf{x}}}{dE_\mathbf{x}}\nabla_{\mathbf{x}}{E}
)
\]
%}
where q is the proposal distribution. When a proposal is accepted, the entropy $S$ of the corresponding output value is updated using the modification factor $f_m$. Otherwise the $S$ of the ``old'' output value will be updated. This sampling procedure repeats until the histogram $H$ passes the flatness check. The sampler then enters the next iteration with the counters in $H$ reset to 0, $\ln f_m$ reduced by half, but the $S$ histogram kept for further accumulation. 
This sampling procedure drives the sampler to visit rare samples whose logit values correspond to the lower entropy, while providing an estimation of entropy $S$ as a result at the end. This proposed algorithm is provided in Alg.~\ref{alg:grad_wl} in Appendix.

\section{Related Works and Discussions}

\noindent\textbf{Performance Characterization} has long been explored even before the era of deep learning~\citep{haralick1992performance,klette2000performance,thacker2008performance}.
The input-output relationship has been explored for simple functions~\citep{hammitt1995determining} and mathematical morphological operators~\citep{gao2002statistical,kanungo1990character}. 
% While there are numerous different approaches for performance characterization~\citep{ramesh1997computer, bowyer1998empirical,aghdasi1994digitization,ramesh1992random, ramesh1994methodology}, 
Compared to existing performance characterization approaches~\citep{ramesh1997computer, bowyer1998empirical,aghdasi1994digitization,ramesh1992random, ramesh1994methodology}, 
% our setting cares the end-to-end input to output distribution~\citep{greiffenhagen2001design} of a model in the entire input space (not task specific) in the blackbox approach~\citep{courtney1997algorithmic,cho1997performance} where the system transfer function from input to output is unknown. 
our work focuses on the output distribution~\citep{greiffenhagen2001design} of a neural network over the entire input space (i.e., not task specific) following the blackbox approach~\citep{courtney1997algorithmic,cho1997performance} where the system transfer function from input to output is unknown. 
% The entire input space setting is thus can be the most general forward uncertainty quantification case~\citep{lee2009comparative} where the model performance is characterized when the inputs are perturbed~\citep{roberts2021principles}. 
Our setting shall be viewed as the most general forward uncertainty quantification case~\citep{lee2009comparative} where the model performance is characterized when the inputs are perturbed~\citep{roberts2021principles}. 
To our best knowledge, we demonstrate for the first time that the challenging task of sampling the entire input space for modern neural networks is feasible and efficient by drawing the connection between neural network and physics models.
% though no concrete performance characterization algorithm or new evaluation metric is proposed. 
% The samples through sampled from our proposed method can be further processed through the performance characterization methods mentioned above.
Our proposed method can offer samples to be further integrated with the performance characterization methods mentioned above.
\vspace{-3mm}
\paragraph{Density Estimation and Energy Landscape Mapping} %\jingbo{please move this to the beginning of the related work. it seems like the most related.}
Previous works in density estimation focus on data density~\cite{tabak2013family??, liu2021density??}, where class samples are given and the goal is to estimate the density of samples. Here we are not interested in the density of the given dataset, but the density of all the valid samples in the pixel space for a trained model. \cite{hill2019building,barbu2020mapping} have done the pioneering work in sampling the energy landscape for energy-based models. Their methods specifically focus on the local minimum and barriers of the energy landscape. We can relax the requirement and generalize the mapping on the ``output'' space where either sufficiently positive or sufficiently negative output (logit) values are meaningful in binary classifiers and other models.

\vspace{-3mm}
\paragraph{Open-world Model Evaluation} Though many neural models have achieved the SOTA performance, most of them are only on in-distribution test sets~\citep{dosovitskiy2020vit, tolstikhin2021mixer, steiner2021augreg, chen2021outperform,zhuang2022gsam, He2015,simonyan2014very, szegedy2015going, huang2017densely, zagoruyko2016wide}. Open-world settings where the test set distribution differs from the in-distribution training set create special challenges for the model. While the models have to detect the OOD samples from in-distribution samples~\citep{liu2020energy,hendrycks2016baseline, hendrycks2018deep,hsu2020generalized,lee2017training,lee2018simple,liang2018enhancing, mohseni2020self, ren2019likelihood}, we also expect sometimes the model could generalize what it learns to OOD datasets~\citep{cao2022openworld, sun2022open}. It has been discovered that models have over-confident predictions for some OOD samples that obviously do not align with human judgments~\citep{nguyen2015deep}. The OOD generalization becomes more challenging because of this discovery, because the models may not be as reliable as we thought they were. Adversarial test sets~\cite{szegedy2013intriguing,rozsa2016adversarial,miyato2018virtual,kurakin2016adversarial,xie2019improving,madry2017towards} also present special challenges as models decisions are different from those of humans. Having a full view of input-output relation with all the above different kinds of test sets under consideration is important.

% \jingbo{I think we don't have to discuss adversarial attacks here. We can just mention it in the experiment result analysis/discussion.}
% \paragraph{Adversarial Attacks} \wei{I don't know if we should mention this because we did NOT observe any adversarial samples during our sampling procedure, not even in our toy example. So I simply add my ideas here without adding the citations. If this should be left, I will add the citation for adversrial attacks as well.} \jingbo{let's hold on for this part. I also feel it's not that related compared with other parts.}
% Newer and sophisticated attacks are developed and brought huge challenges to machine learning models when deployed. Ideally, with the full picture of input-output statistic as we propose, we should be able to pick a either peaks with high or low logit values and observe the adversarial samples. Unfortunately, we do not observe them in our experiments. We suspect this is because these samples  . Previous works only demonstrate the adversarial samples exist, but whether they dominant the high/low logits regions remains a unanswered question. If our sampler is correct, this . No matter what, getting a full picture of the input-output relationship should be able to conclusively tell how serious the adversarial attack problem is for a specific model or a type of models and whether the smartly developed defense algorithm really work in those models. 
\vspace{-3mm}
\paragraph{Samplers} 
% \wei{modified} 
MCMC samplers~\citep{chen2014stochastic, welling2011bayesian, li2016preconditioned, xu2018global} are developed to scale to big datasets and sample efficient with gradients. Recently, Gibbs-With-Gradients (GWG)~\citep{grathwohl2021oops} is proposed to pick the promising pixel(s) as the proposal.
To further improve sampling efficiency, CSGLD~\citep{CSGLD} drives the sampler to explore the under-explored energy using  similar idea as Wang--Landau algorithm~\citep{wang2001efficient}. 
The important difference between our problem setting and the previous ones solved by other MCMC samplers is the function or model as distribution to be sampled from is unknown. Wang--Landau algorithm utilizes previous approximation of the distribution to drive the sampler to explore the under-explored energy regions. This algorithm can be more efficient through parallelization~\citep{vogel2013generic, cunha2008improving}, assumption about continuity in output space \citep{junghans2014molecular,li2017histogram} and extension to multi-dimensional outputs~\citep{PhysRevLett.96.120201}. While the previous samplers can be applied to high-dimensional inputs, the energy functions in physics are relative simple and symmetric. However, modern neural networks are complex and hard to characterize performance~\citep{roberts2021principles}. We assume agnostic of the output properties of the model and thus apply the Wang--Landau algorithm to sample the entropy as a function of energy but with the gradient proposal in GWG to make the sampler more efficient. Similar to GWG, our sampler can propose the inputs corresponding to the under-explored regions of outputs. Improvements of efficiency can benefit from a patch of pixel changes. 

\section{Experiments~\label{sec:exp}}

In this section, we apply our proposed Gradient Wang--Landau sampler to inspect a few neural network models and present the discovered output histogram together with representative samples.
The dataset and model training details are introduced in Sec.~\ref{sec:data}.
We first empirically confirm our sampler performance through a toy example in Sec.~\ref{sec:toy}.
We then discuss results for modern binary classifiers in Sec.~\ref{sec:real} and Sec.~\ref{sec:real_res}. Hyperparameters of the samplers tested in are Appendix~\ref{app:sampler}.

\subsection{Datasets, Models, and Other Experiment Settings~\label{sec:data}}
\vspace{-3mm}
\paragraph{Datasets}
As aforementioned, we focus on binary classification. 
Therefore, we derive two datasets from the MNIST datasets by only including samples with labels $\{0, 1\}$. The training and test splits are the same as those in the original MNIST dataset.  
\begin{itemize} [nosep,leftmargin=*]
    \item \textbf{Toy} is a simple dataset with $5\times5$ binary input images we construct. It is designed to make feasible the bruteforce enumeration over the entire input space (only $2^{5\times5}$ different samples). 
            We center crop the MNIST samples from $\{0, 1\}$ classes and resize them to $5\times5$ images. 
            We compute the average of the pixel values and use the average as the threshold to binarize the images --- the pixel value lower than this threshold becomes $0$; otherwise, it becomes $1$. The duplicates are not removed for accuracy after resizing since PyTorch does not find duplicate row indices.
            %\jingbo{is there any duplicates after resizing and binarization? how did we deal with this? provide some counts on the training, validation, and test set sizes will be helpful.}
    \item \textbf{MNIST-0/1} is an MNIST dataset whose samples only have the {0,1} labels. To align with the GWG setting, the inputs are discrete and not Z-normalized. 
            Therefore, in this dataset, the input $\mathbf{x}$ is $28\times28$ dimensional with discrete pixel values from $\{0,...,255\}$. 
            
\end{itemize}
%\jingbo{since we mentioned test/validation set in the compared methods, please also talk about the train/validation/test splits here. If we follow the standard split, we can just say it.}
\vspace{-3mm}
\paragraph{Neural Network Models for Evaluation}
Since the focus of this paper is not to compare different neural architectures, given the relatively small datasets we have, we train two types of models, a simple CNN and \textbf{ResNet-18}~\citep{He2015}. Each pixel of the inputs is first transformed to the one-hot encoding and passed to a 3-by-3 convolution layer with 3 channel output.
The \textbf{CNN} model contains 2 convolution layers with 3-by-3 filter size.
The output channels are 32 and 128. 
The final features are average-pooled and passed to a fully-connected layer for the binary classification.
% Both the ResNet-18 and the CNN are tuned so that they can achieve almost perfect in-distribution test set accuracy (i.e., over 97\%) as shown in each the following sections.

Please keep in mind that our goal in this experiment section is to showcase that our proposed sampler can uncover some novel interesting empirical insights for neural network models. 
Models with different architectures, weights due to different initialization, optimization, and/or datasets will lead to different results.
Therefore, our results and discussions are all \emph{model-specific}.
Specifically, we train a simple CNN model to classify the $5\times5$ binary images in the Toy dataset (\textbf{CNN-Toy}). 
The test accuracy of this CNN-Toy model reaches $99.7\%$, which is almost perfect. 
We train a simple CNN model to classify the $28\times28$ grey-scale images in the MNIST-0/1 dataset (\textbf{CNN-MNIST-0/1}).
The test accuracy of CNN-MNIST-0/1 model is $97.8\%$.
We train a ResNet-18 model to classify the $28\times28$ grey-scale images in the MNIST-0/1 dataset (\textbf{ResNet-18-MNIST-0/1}).
The test accuracy of ResNet-18-MNIST-0/1 model is $100\%$.

\vspace{-3mm}
\paragraph{Sampling Methods for Comparison}
We compare several different sampling methods (including our proposed method) to obtain the output histogram over the entire input space. 
\begin{itemize}[nosep,leftmargin=*]
    \item \textbf{Enumeration} generates the histogram by enumerating all the possible pixel values as inputs. This is a rather slow but the most accurate method.
    \item \textbf{In-dist Test Samples} generates the histogram of the inputs based on the fixed test set.% \jingbo{make it clear, is it validatation or test?}. 
    This is commonly used in machine learning evaluation. It is based on a very small and potentially biased subset of the entire input space.  
    \item Wang-Landau algorithm (\textbf{WL}) generates the histogram the Wang-Landau algorithm with the random proposal. Specifically, we randomly pick one pixel at a time and change it to any valid (discrete) value as in this implementation~\footnote{\url{https://www.physics.rutgers.edu/~haule/681/src_MC/python_codes/wangLand.py}}. 
    \item Gradient Wang-Landau (\textbf{GWL}) generates the histogram by our proposed sampler of Wang-Landau algorithm with gradient proposal.  
\end{itemize}

\begin{figure}[t]
    \centering
    \includegraphics[width=1.0\linewidth]{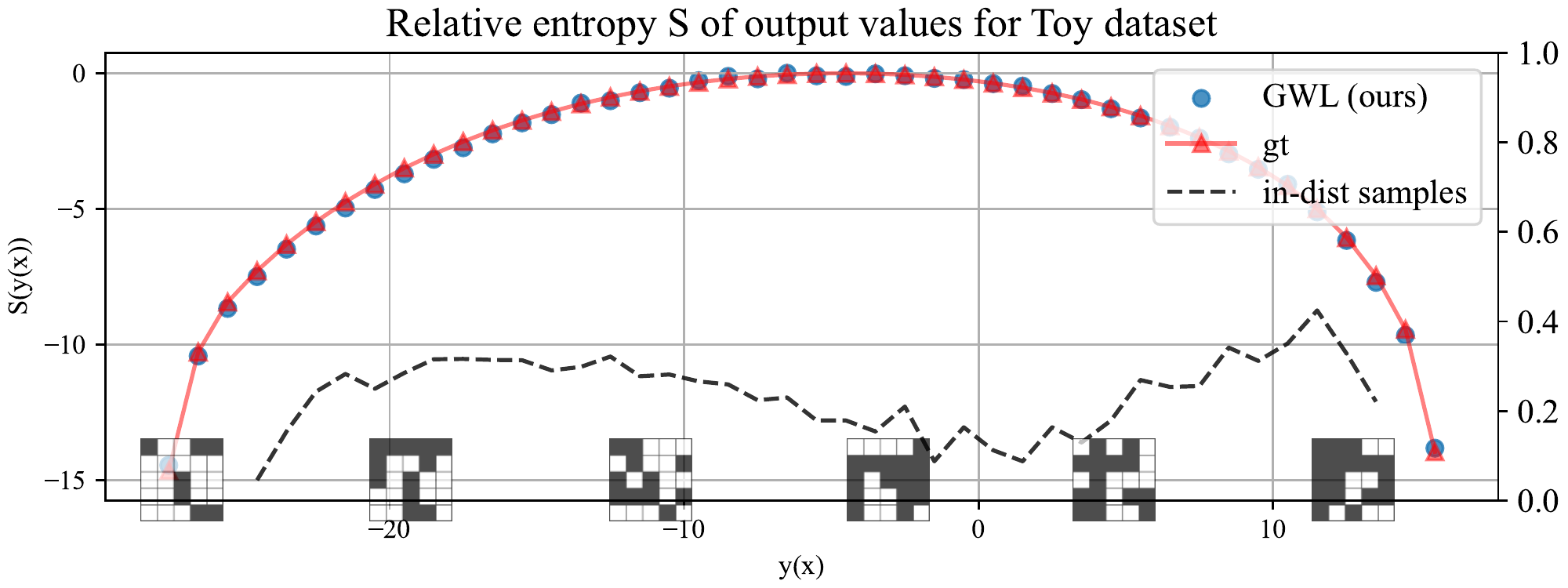}
    \vspace{-3mm}
    \caption{Output histograms of CNN-Toy obtained by different sampling methods.  The in-distribution samples are only a very small portion in the output histogram. 
    We also present the representative samples obtained by GWL given different logit values.}
    \label{fig:toy}
    \vspace{-3mm}
\end{figure}

\subsection{Results of CNN-Toy~\label{sec:toy}}

Given the CNN-Toy model, we apply Enumeration, GWL, and In-dist Test Samples to obtain the output entropy histograms, as shown in Fig.~\ref{fig:toy}.
Note that our GWL method samples the relative entropy of different energy values as duplicate $\mathbf{x}$ may be proposed. 
After normalization with the maximum entropy, the GWL histogram almost exactly matches the Enumeration histogram which is the ground truth histogram.
This confirms the accuracy of our GWL sampler and we can apply it further to more complicated models with confidence.

Remarkably, this histogram is quite different from the expectation we presented in Fig.~\ref{fig:intro2} --- this histogram is even not centered at $0$ or has the expected subdominant peaks on both the positive and negative sides. 
Instead, the dominant peak is so wide that it covers almost the entire spectrum of the possible output values. 
From a coarse-grained overview, most of the samples are mapped to the center of logit $-5$ with a decay from $-5$ to both sides in the CNN-Toy model. 
This shows the CNN-Toy model is biased to predict more samples to the negative logit values.

In Fig.~\ref{fig:toy}, we also present the representative samples obtained by GWL given different logit values in the CNN-Toy model. Our conjectured analysis of the representative samples are in Appendix~\ref{app:toy_representative}.
From this example, one can see that the output histogram over the entire input space can offer a comprehensive understanding of the neural network models, helping researchers better understand critical questions such as the distribution of the outputs, where the model maps the samples to, and what the representative samples with high likelihood are.

% \begin{figure}[t]
%     \centering
%     \includegraphics[width=8cm]{fig/final_cnn.pdf}
%     \vspace{-3mm}
%     \caption{Output histograms of CNN-MNIST-0/1 obtained by different sampling methods. 
%         The blue scale is for GWL and the red scale is for In-distribution Test Samples. 
%         We also present the representative samples obtained by GWL given different logit values (more in Fig.~\ref{fig:cnn_imgs} in Appendix) \wei{this plot needs to be updated} }
%     \vspace{-3mm}
%     \label{fig:cnn_final}
% \end{figure}

\begin{figure}[t]
    \centering
         \includegraphics[width=0.99\linewidth]{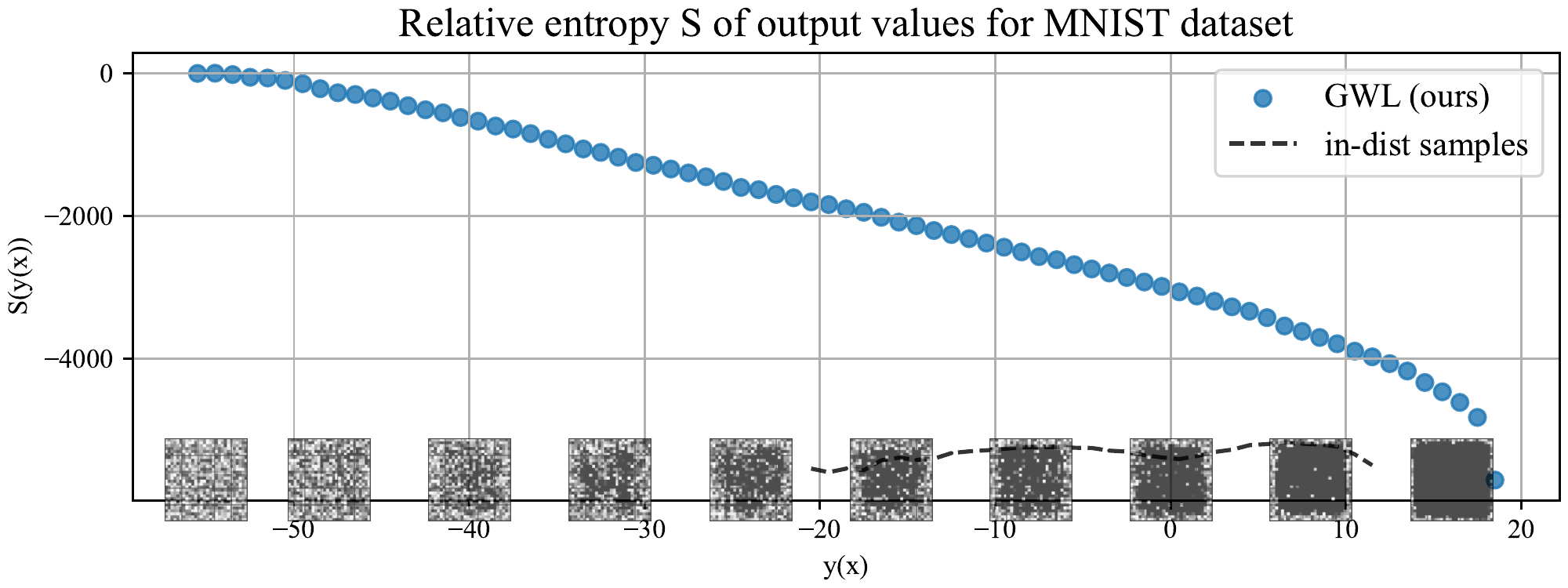}
         \vspace{-3mm}
         \label{fig:final_cnn}
     % \subfigure[Results with test set re-initialization.]{
     %     \centering
     %     \includegraphics[width=0.45\linewidth]{fig/final_cnn.pdf}
     %     \vspace{-3mm}
     %     %\label{fig:final_res2}
     % }
    \caption{Output histograms of CNN-MNIST-0/1 obtained by different sampling methods. 
        The blue scale is for GWL and the black scale is for in-distribution test samples. 
        We also present the representative samples obtained by GWL given different logit values (more in Fig.~\ref{fig:cnn_imgs} in Appendix). }
    \vspace{-3mm}
    \label{fig:cnn_final}
\end{figure}

\subsection{Results of CNN-MNIST-0/1~\label{sec:real}}
\paragraph{Entropy Histogram from GWL} The application of GWL on the CNN-Toy model is encouraging. Now we apply GWL to the CNN-MNIST-0/1 that is trained on a real-world dataset.
The results from the $5^\textrm{th}$ iteration are shown in Fig.~\ref{fig:cnn_final}. 
As our GWL reveals, the output histogram of CNN-MNIST-0/1, similar to CNN-Toy's histogram, does not have the subdominant peaks. It is also different from the presumed case in Fig.~\ref{fig:intro2}.
Compared with the output histogram of the CNN-Toy model (i.e., Fig.~\ref{fig:toy}), for the CNN-MNIST-0/1 case, the peak is on the negative boundary and the histogram is skewed towards the negative logit values. 
$S$ monotonically decreases as the logit values go from negative to positive. 
While the in-distribution samples have logit values between $-20$ and $12$ as we expect, these samples are exponentially (i.e., $e^{2000}$ at logit value -20 to $e^{5500}$ at logit value 18, thousands in log scale) less often found than the majority samples whose logit values are around $-55$.
From a fine-grained view, the CNN-MNIST-0/1 model tends to map the human-unrecognizable samples to the very negative logit values. 
While previous work~\citep{nguyen2015deep} showed the existence of the overconfident prediction samples, our result shows a rough but quantitative performance of this CNN which can serve as a baseline for further improvements.

\begin{figure*}
     \centering
     \subfigure[$S$ of GWL for steps.]{
         \centering
         \includegraphics[width=0.31\linewidth]{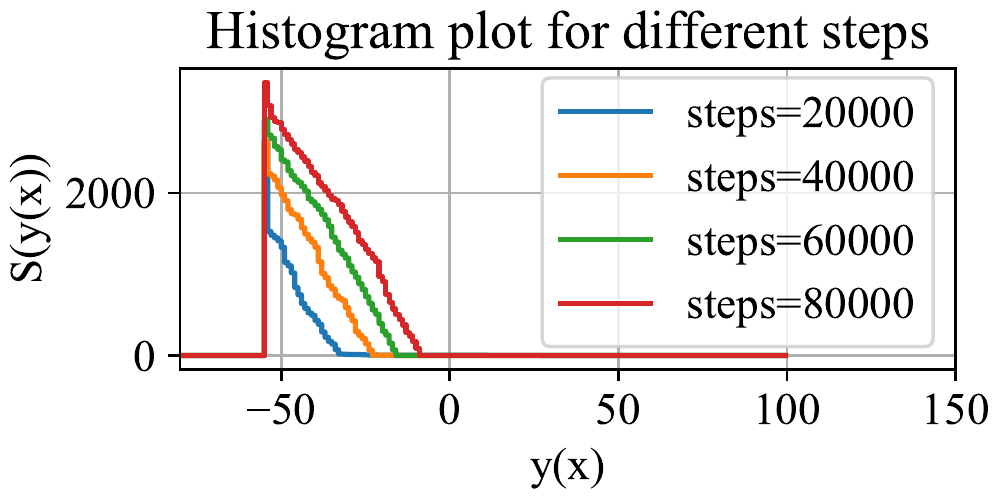}
         \label{fig:first_iter}
     }
     \subfigure[$S$ of GWL per iteration.]{
         \centering
         \includegraphics[width=0.31\textwidth]{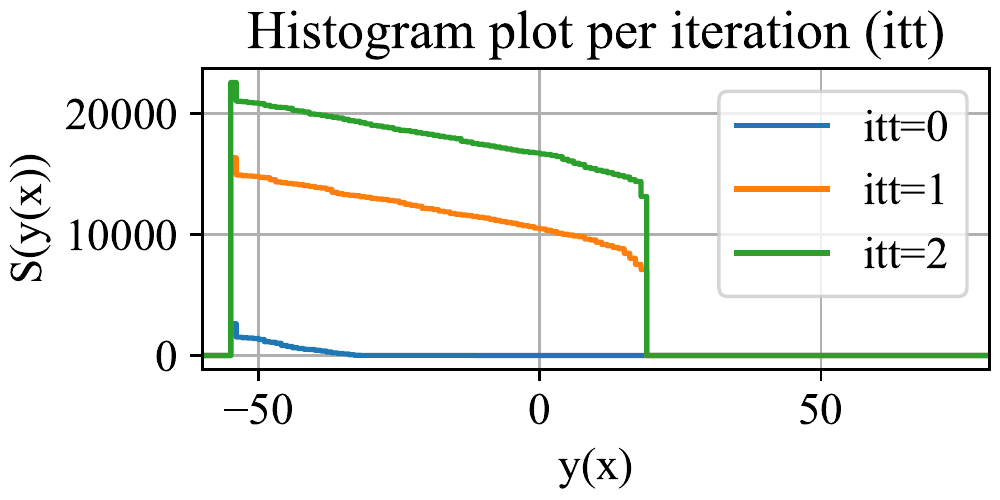}
         \label{fig:all_iter_gwl}
     }
     \subfigure[$S$ of WL per iteration.]{
         \centering
         \includegraphics[width=0.31\linewidth]{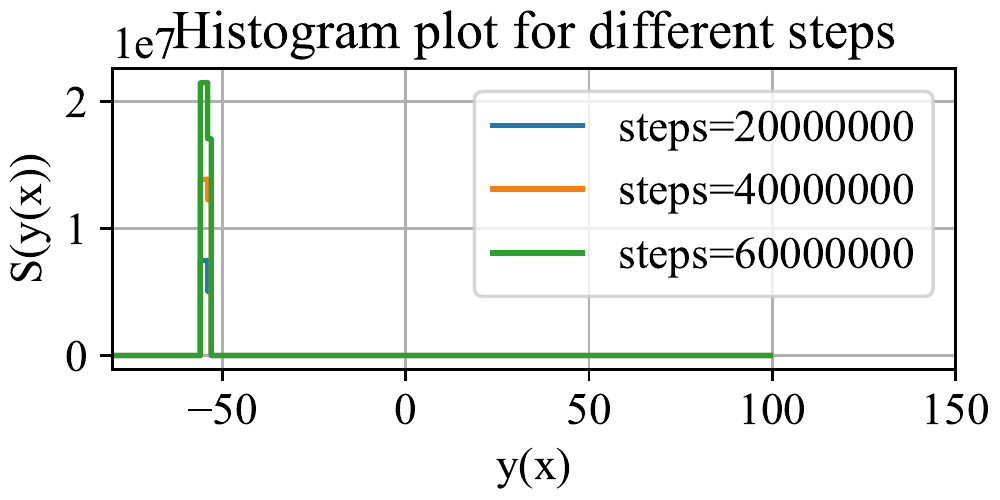}
         \label{fig:all_iter_wl}
    }
    %\vspace{-3mm}
    \caption{Intermediate output histogram $S$ per iteration. (a) GWL gradually explores the logit values in the first iteration. (b) GWL discovers the output histogram well within 2 iterations. (c) The original WL explores the output distribution much slower.}
    \label{fig:intermediate}
    \vspace{-3mm}
\end{figure*}

%While previous work~\citep{nguyen2015deep} showed the existence of the overconfident prediction samples, we know in Fig~\ref{fig:cnn_final} the relative counts between them vs the expected in-distribution logit regions for CNN: the most dominant region is around 3100 (logit around -55) verses 1 (logit around 13) in log scale and the overconfident regions (logit around -55) do not correspond to samples with $0$ class. This is a rough but quantitative performance of this CNN which can serve as a baseline for further improvements. 
\vspace{-3mm}
\paragraph{GWL is much more efficient than WL}
We first confirm the correctness of our WL sampler on a $16 \times 16$ Ising model and apply it to this CNN model. WL takes a much longer time to converge and we are not able to obtain the converged results. Both WL and GWL cannot have more than 1 worker writing to the same set of DOS bins or else incorrect DOS will be resulted~\cite{yin2012massively}.
For comparison, we inspect the intermediate $S$ results of the GWL and WL samplers, as shown in Fig.~\ref{fig:intermediate}.
As one can see from Fig.~\ref{fig:first_iter}, GWL is already able to explore the logit values efficiently from the most dominant output value around $-55$ to the positive logit values in the first iteration. 
Within only two iterations (Fig.~\ref{fig:all_iter_gwl}), GWL can discover the output histogram covering the value range from $-55$ to $18$. 
On the other hand, as presented in Fig.~\ref{fig:all_iter_wl}, the original WL can only explore the output ranges from around $-55$ to $-53$ for 60,000,000 steps (around 10 days without much substantial progress). WL converges significantly slower and never ends in a reasonable time. 
This result indicates that the GWL converges much faster than the original WL and is able to explore a much wider range of output values. 

\begin{figure*}
     \centering
     \subfigure[Output distribution with logit values from -300 to -30.]{
         \centering
         \includegraphics[width=0.45\linewidth]{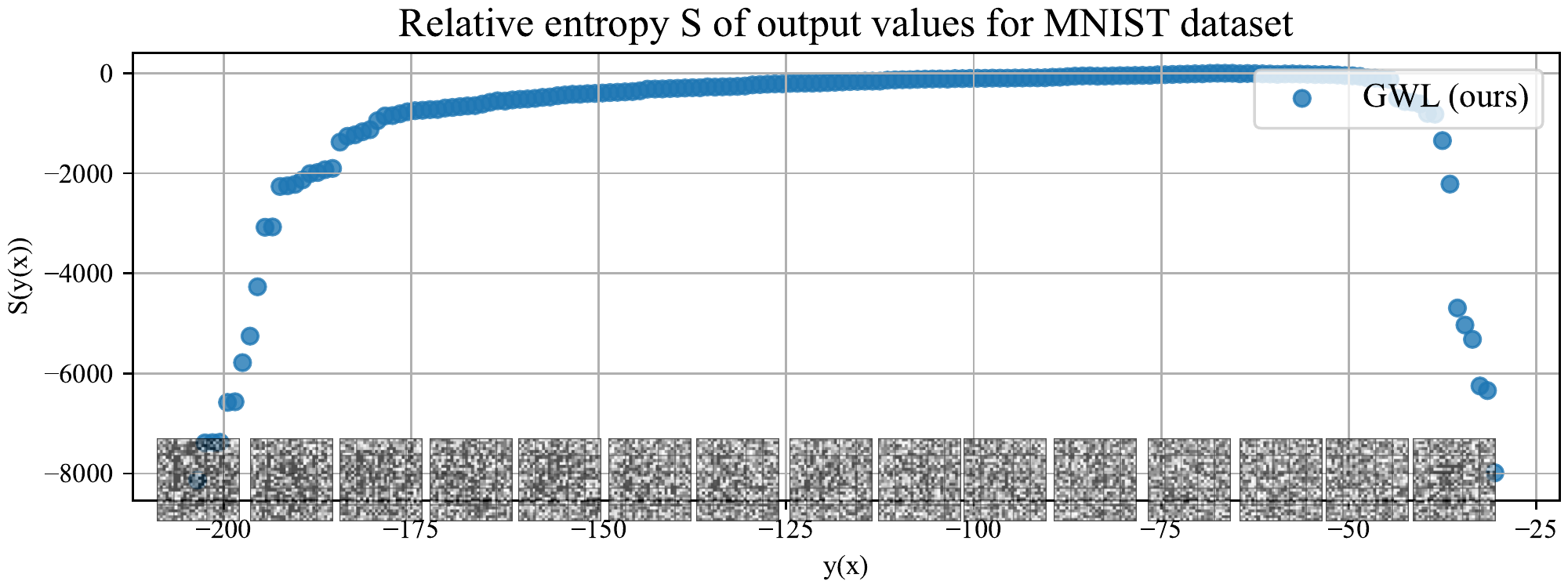}
         \vspace{-3mm}
         \label{fig:final_res1}
     }
     \subfigure[Output distribution with logit values from -30 to 20.]{
         \centering
         \includegraphics[width=0.45\linewidth]{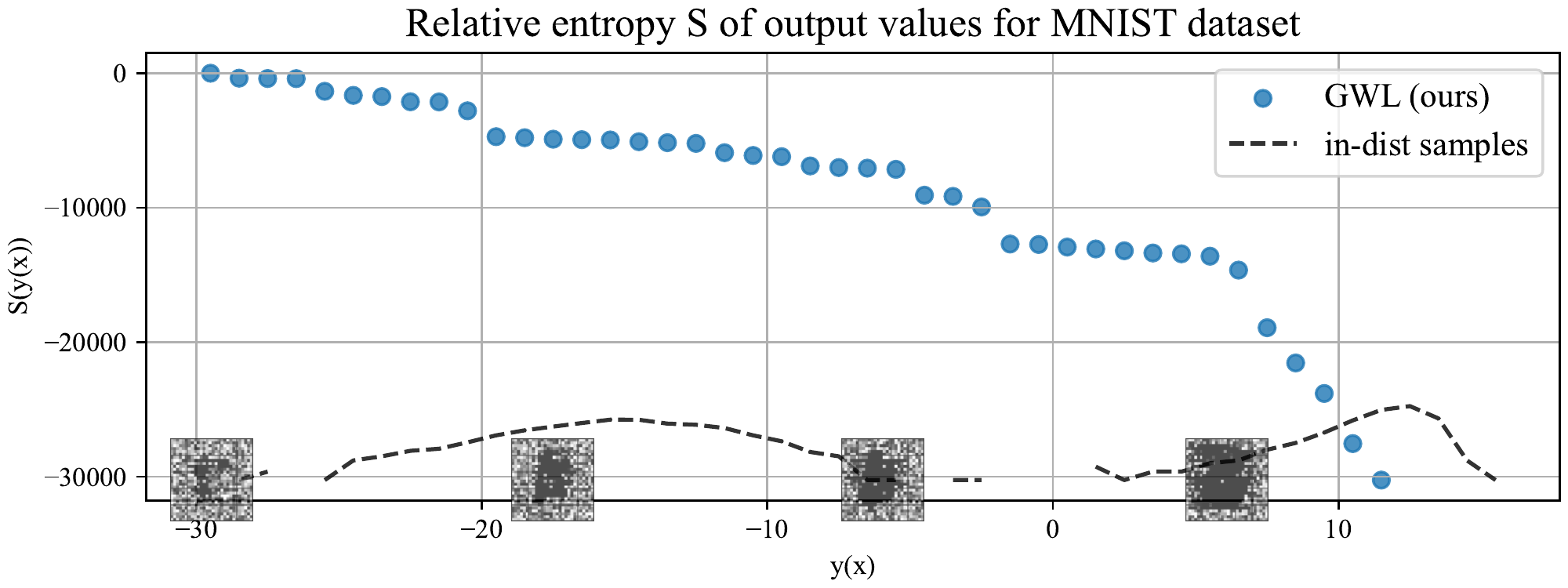}
         \vspace{-3mm}
         \label{fig:final_res2}
     }
    %\vspace{-3mm}
    \caption{Output histograms of ResNet-18-MNIST-0/1 obtained by different sampling methods. 
        There may be a sharp local minima in the output landscape causing a cliff around the logit value of -30.
        The blue scale is for GWL and the black scale is for in-distribution test samples. 
        We also present the representative samples obtained by GWL given different logit values. (more in Fig.~\ref{fig:res_imgs} in Appendix)
        }
    %\vspace{-3mm}
    \label{fig:res_final}
\end{figure*}

\vspace{-3mm}
\paragraph{Manual inspection on more representative samples~\label{para:manual}} 
As show in Fig.~\ref{fig:cnn_final}, for the CNN-MNIST-0/1 model, GWL can effectively sample input images from logit values ranging from -55 to 18. 
We further group these logit values per 5 unit of logit value in $S$. %\jingbo{not sure how to end up with the 10 groups. may want to make it clear about the range of every group.}
For every group, we sample 200 representative input images.
To make sure they are not correlated, we sample every 50000 pixel changes. 
For demonstration purposes, we randomly pick 10-out-of-200 samples from every group in Fig.~\ref{fig:cnn_imgs} in Appendix.
We manually inspect the sufficiently positive group (e.g., the last column in Fig.~\ref{fig:cnn_imgs}) and the sufficiently negative groups (e.g., the first five columns in Fig.~\ref{fig:cnn_imgs}) , and there are no human recognizable samples of digits.
We also observe an interesting pattern that as the logit value increases, more and more representative samples have black background.
This result suggests that the CNN-MNIST-0/1 model may heavily rely on the background to classify the images~\citep{xiao2020noise}. We conjecture that is because the samples in the most dominant peak are closer to class 0 samples than class 1 samples and this is supported by experimental results (see Appendix.~\ref{app:samples}).
More rigorous experiments to a definite conclusion is yet required as future work.
In summary, although CNN-MNIST-0/1 holds a very high in-distribution test accuracy, it is far from a robust model because it does not truly understand the semantic structure of the digits.  

\vspace{-3mm}
\paragraph{Discussion} Fig.~\ref{fig:cnn_final} presents challenges to the OOD detection methods that may be more model-dependent than we thought before. If the model cannot map most of the human unrecognizable samples with high uncertainty, the likelihood-based OOD detection methods~\citep{liu2020energy,hendrycks2016baseline} cannot perform well for samples in the entire input space. Fig.~\ref{fig:cnn_imgs} shows the inputs with the in-distribution output values (output logits of the red plot) of the CNN model may not uniquely correspond to in-distribution samples. More rigorous experiments to a definite conclusion are yet required as future work. 

\vspace{-3mm}
\subsection{Results of ResNet-18-MNIST-0/1~\label{sec:real_res}}
\paragraph{Entropy Histogram from GWL} When applying our GWL samplers to the ResNet-18-MNIST-0/1 model, for $0^{\textrm{th}}$ iteration (Fig.~\ref{fig:final_res1}), we observe that the sampler discovers a wide range of negative logit values from around logit value of -220 to around -33, much wider than that of the CNN's. 
This range of negative logits, however, does not correspond to human recognizable inputs and there is no obvious pattern observed in contrast to CNN-MNIST-0/1's results. It means the ResNet-18-MNIST-0/1 model makes more confident predictions for some samples than the CNN-MNIST-0/1 model does. Moreover, we observe a cliff around the logit value of -33 and thus we specifically sample the region from -30 to 20 and generate the representative samples in this region where the in-distribution logits fall into. Fig.~\ref{fig:final_res2} shows the entropy histogram after the $1^{\textrm{st}}$ iteration. Some output regions of the in-distribution samples take longer time to discover. This calls for a more efficient sampler in the future. 

\vspace{-3mm}
\paragraph{Manual inspection on more representative samples} Interestingly, similar (if not exactly the same) pixel patterns for CNN-MNIST-0/1 model appear, as shown in Fig.~\ref{fig:final_res2} and Fig.~\ref{fig:res_imgs}. The representative samples, however, have broader noisy boundaries compared to those from the CNN-MNIST-0/1 model. The same phenomenon also happens that the double peaks of the test set samples do not align with the output distribution of the entire input space.   

Because of the complexity of ResNet-18 over CNN and it takes a longer time to converge, we do not draw conclusions about ResNet-18-MNIST-0/1 evaluation of the entropy difference.
%Our primitive results indicate the representative samples of the ResNet-18-MNIST-0/1 models have human recognizable structures. 
Compared with the CNN-MNIST-0/1 model, ResNet-18-MNIST-0/1 has more interesting phenomena for further exploration.

\section{Conclusion}
We aim to get a full picture of the input-output relationship of a model through the inputs valid in the pixel space. We propose to obtain a histogram to estimate the entropy in the output space to better understand the input-output distribution. When the inputs are high-dimensional, enumeration or uniform sampling is either impossible or takes too long to converge. We connect the density of states in physics to this histogram of output entropy. We propose a new, efficient sampler, Wang--Landau sampling with gradient proposals, to achieve this goal. We confirm empirically this can be achieved and uncover some new aspects of neural networks.

We observe several limitations. First, though we combine two samplers that have the theoretical guarantee of convergence and confirm the performance of the sampler through empirical results, we do not provide a proof of convergence when they are combined. Second, because of the nature of our problem, we observe that the sampler still takes a decent amount of time to converge, especially for the more complicated network architectures such as ResNet. We avoid making conclusions on the distributions but provide some observations for ResNet. The sampler for ResNet is still converging but it also calls for further development of faster samplers for these more complicated networks. Third, even though the ratio of the recognizable samples can be derived from our sampler, our CNN model maps an enormous amount of samples to the desired output region of the in-distribution inputs, and we do not observe even one human recognizable sample out of the hundreds of representative samples. Future automatic methods can alleviate the need of human labels.

% First, we do not observe adversarial samples in this output distribution. Adversarial samples seem not dominate the output distribution. 
% $S$ and its representative samples show the models may be fundamentally subject to overconfident prediction. Based on our results, the likelihood based OOD detection methods may not work without fine-tuning~\cite{liu2020energy}. 

For future work, it is necessary to develop new and more efficient samplers that have theoretical guarantees to acquire this input-output relationship in order to sample with more pixels, such as the ImageNet~\citep{5206848}. Most importantly, we can then develop new insights into network architectures developed in the last decade for \emph{open-world} applications using these efficient samplers. %A full picture of model output behavior through all the possible input samples is the new gate for a comprehensive understanding of neural networks.  

%\section{Reproducibility}
%We provide fairly amount of information to re-implement our sampler. The data processing is in the Sec.~\ref{sec:data} and algorithm is in Appendix~\ref{app:algorithm}. The hyperparameters and sampling details are also listed in the Sec.~\ref{sec:exp}. We also provide different time stamp of steps for our samplers to indicate what to expect during the sampling procedure in Fig.~\ref{fig:intermediate}. Of course, the Wang-Landau algorithm we adopted is the prototypical one and it subjects to some issues reported in its follow-up works, such as the discontinuity of the boundaries between bins and trapping in one of the bins. These problems lead to some issues in our experiments and we discussed them in Sec.~\ref{sec:real_res}. More advanced algorithms have been developed to resolve these issues.

%\section{Ethics Statement}
%Our method aims to provide a comprehensitve understanding of the neural models. This work will be applicable to many applications, such as those in the safety and trusty-worthy machine learning. As a pilor study, we do not anticipate the negative aspects of our work. 


%\bibliography{example_paper}
\bibliographystyle{icml2023}

%%%%%%%%%%%%%%%%%%%%%%%%%%%%%%%%%%%%%%%%%%%%%%%%%%%%%%%%%%%%%%%%%%%%%%%%%%%%%%%
%%%%%%%%%%%%%%%%%%%%%%%%%%%%%%%%%%%%%%%%%%%%%%%%%%%%%%%%%%%%%%%%%%%%%%%%%%%%%%%
% APPENDIX
%%%%%%%%%%%%%%%%%%%%%%%%%%%%%%%%%%%%%%%%%%%%%%%%%%%%%%%%%%%%%%%%%%%%%%%%%%%%%%%
%%%%%%%%%%%%%%%%%%%%%%%%%%%%%%%%%%%%%%%%%%%%%%%%%%%%%%%%%%%%%%%%%%%%%%%%%%%%%%%
\newpage
\appendix
\onecolumn
\section{Gradient Wang-Landau Algorithm~\label{app:algorithm}}
Here we provide the algorithms of the GWL algorithm. The input and output are listed. The hyperparameters are determined mostly by the toy-example. 

\begin{algorithm}[h]
\caption{Our proposed Gradient Wang-Landau (\textbf{GWL})}
\label{alg:grad_wl}
\begin{algorithmic}
\STATE {\bfseries Input:} pretrained model $y$: $\mathbf{x} \rightarrow z$, flat histogram $H=0$, entropy histogram $S=0$, modification factor $\ln f_m$,
% \jingbo{this lnf looks like infinity... is this some tech term popular in W-L?} \wei{yes, this is poplular in W-L. The `ln' is the reads as natural log. The idea is instead of adding 1 as count, we add 1 in the log scale so that it can converge faster. }, 
number of iterations T, test set $\mathcal{D}_{te}$, GWG sampler $GWG(z,S)$, interpolation function $g(z,S)$ 
% \jingbo{it looks like we never update $g(y, S)$ from the algorithm? a little counter-intuitive to me.}\wei{The linear or cubic interpolation is parameter free and it only depends the histogram.}
\FOR{$i=1$ {\bfseries to} $T$}%\jingbo{how to we define the convergence? looking at S? or just lnf?}
    \STATE $\mathbf{x}$ $\sim$ $\mathcal{D}_{te}$
    \REPEAT %{H is not flat}
    % \jingbo{how to define the flat? y is a real number right? do we use bins?}\wei{yes, we do use bins. We also need to round the y to the closest bin y' for one of the bins. The flatness check has different standards. Do we need to specify it?} \jingbo{either mentioned it in full text or here.}}
        \STATE $z$ = $y(\mathbf{x})$
        \STATE $S_{in}$ = $g(z,S)$
        %\Comment{Get the continuous interpolation entropy $S_{in}$ at output $z$}
        \STATE $\mathbf{x}$ $\sim$ $GWG(z, -S_{in})$ 
        %\Comment{Take the negation of $S_{in}$ for Wang-Landau exploration}
        % \jingbo{Where does the potential rejection happen? Is it somewhere inside GWG?} \wei{yes, both the WL and GWG has rejection mechanisms.}
        \STATE $\Tilde{z}$ = round($z$)
        %\Comment{Round $z$ to the nearest $z'$ that corresponds to one of the bins}
        \STATE S[$\Tilde{z}$] $\leftarrow$ S[$\Tilde{z}$]+ $\ln f_m$
        \STATE H[$\Tilde{z}$] $\leftarrow$ H[$\Tilde{z}$]+1
    \UNTIL{H is flat}
    \STATE $\ln f_m$ $\leftarrow$ ($\ln f_m$)/2
    \STATE $H \leftarrow 0$ 
    %\Comment{Reset all the bins in counter $H$ to 0}
    %\jingbo{maybe better to use $\leftarrow$ for those ``assignment'' operators. worth a comment here to state that we are resetting the counter histogram?}
\ENDFOR
\STATE \textbf{return} S($y$) 
% \jingbo{what's the final output? S?}
\end{algorithmic}
\end{algorithm}

\section{Comment on Representative samples for CNN-Toy~\label{app:toy_representative}}
The visualization results suggest that the CNN-Toy model probably learns the digit ``1'' for positive logit values as the center pixels of the representative samples are white (see the three representative samples with logit values from 0 to 20) and ``0'' for the very negative logit values as the center pixels of the representative samples are black (see two representative samples with logit values from -20 to -30).

\section{Hyper-parameters and Implementation Details for GWL and WL~\label{app:sampler}}
The hyper-parameters for GWL and WL are extremely similar, if not identical, as the only major difference between GWL and WL is the gradient proposal versus the random proposal. 
We first preset a large enough range of output values for the sampler to explore the trained neural network models. 
In our experiments, we found that the output (logit) values of the binary classifiers typically fall in the range of -300 to 100 (based on ResNet). %\jingbo{is this range based on Toy dataset?}
Therefore, we use this range for all experiments.
For flatness histogram $H$, the bin window size is set to be 1, resulting in 400 bins.
The histogram $H$ is considered flat if the difference between maximum bin value and minimum bin value is smaller than the average bin value. 
%\wei{For output histogram $S$, we set the bin window size to be 0.1, resulting in 4000 bins.
%Instead of updating one bin at a time for $S$, we update the neighbor bins with exponential decay. 
%We use the linear interpolation to approximate the bins for continuous queries. 
%We iterate 5 times with test set initialization. Every step the GWG tries to at most update 10 pixels.}

\section{Samples similarity~\label{app:samples}}
We conjecture that is because the samples in the most dominant peak are closer to class 0 samples than class 1 samples. We compute the L2 pixel-wise distance from the uniform noise image to the samples of class 1 and 0 respectively. The mean L2 distance from uniform noise to 0 is around 0.3121 and that from uniform noise to 1 is around 0.3236. The distance between 1 and 0 samples is 0.1652. This result shows the samples in the most dominant peak are closer to class 0 samples than class 1 samples. 

\section{Representative inputs~\label{app:repre_imgs}}
Here we list more representative samples of the CNN-MNIST-0/1 scenario. The samples are bounded by a black box of boundaries. 

\begin{figure*}%[t]
     \centering
     \subfigure[CNN-MNIST-0/1]{
         \centering
         \includegraphics[width=0.6\linewidth]{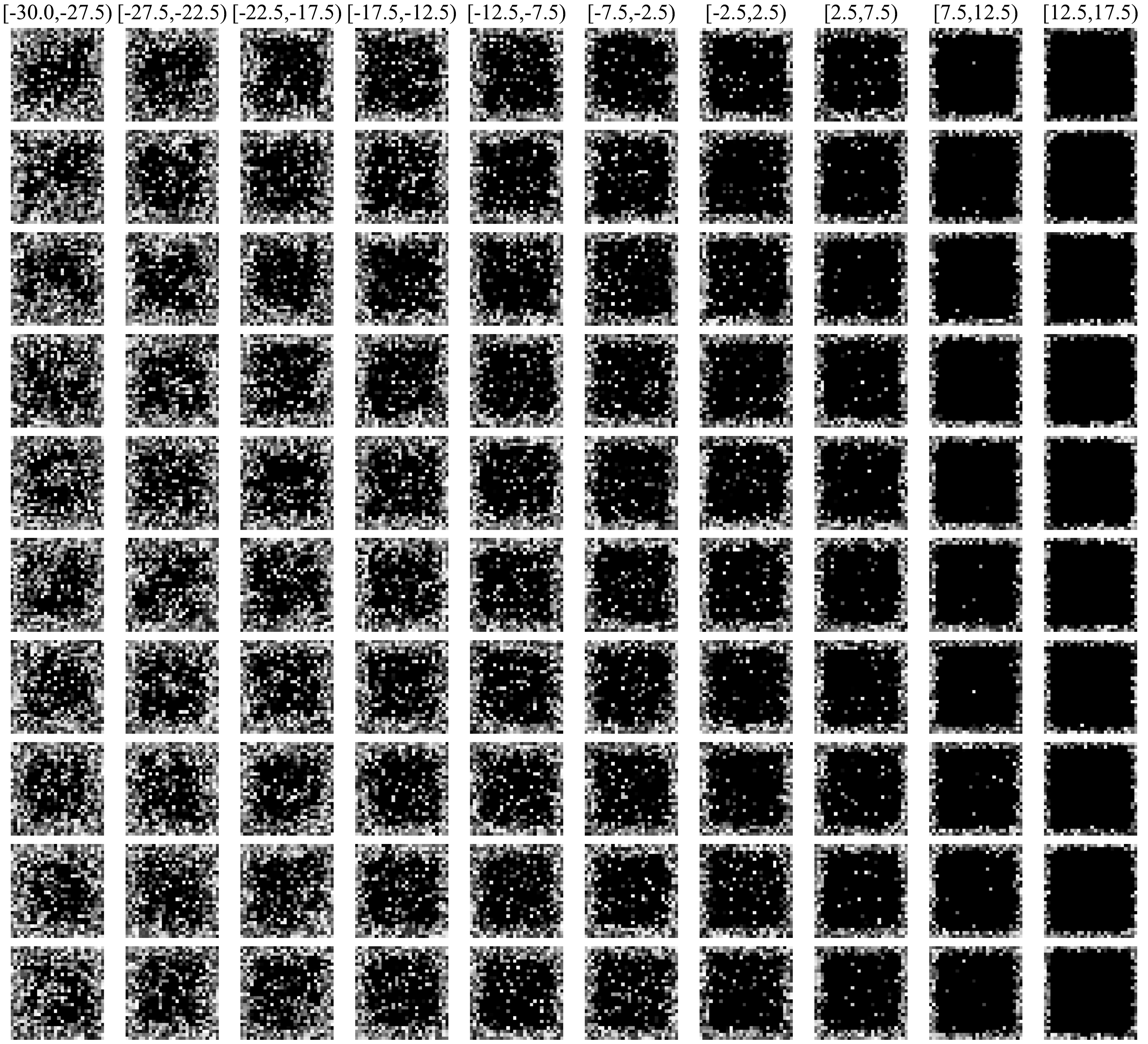}
         %\vspace{-3mm}
         %\caption{Different types of input samples}
         \label{fig:cnn_imgs}
     }
     \subfigure[ResNet-18-MNIST-0/1]{
         \centering
         \includegraphics[width=0.6\linewidth]{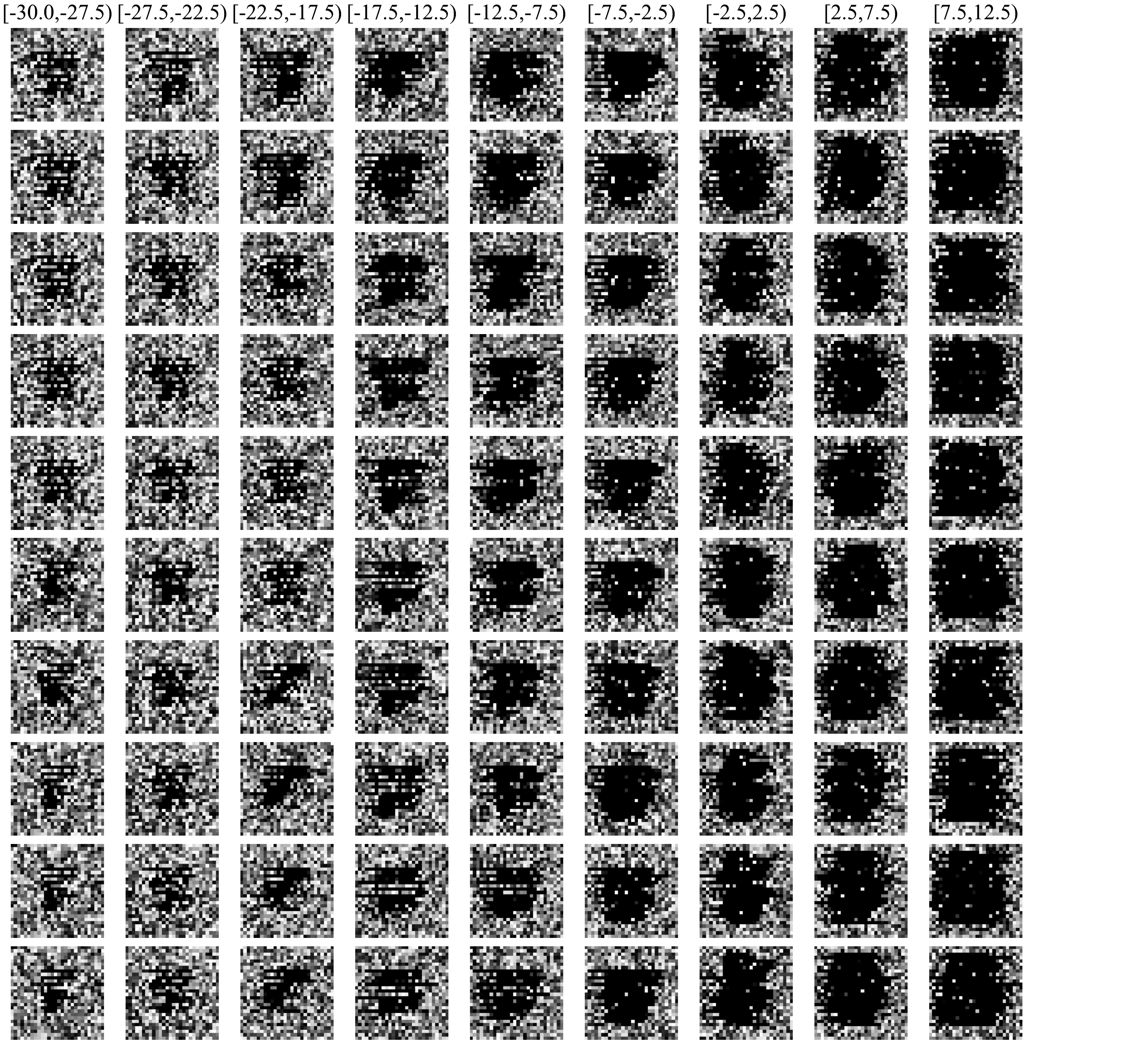}
         %\vspace{-3mm}
         %\caption{The histogram of an example binary classifier}
         \label{fig:res_imgs}
     }
    \caption{More representative samples of the CNN-MNIST-0/1 model and ResNet-18-MNIST-0/1 obtained by GWL at different logit values, grouped by logit values larger than -30.
     We further group these logit values per 5 bins (correspond to a difference of 5 in logit value) in $S$. The output values in the first column are within the range [-30,-27.5) etc.}
\end{figure*}

\end{document}